\documentclass[review, times, 10pt]{elsarticle}

\usepackage{amssymb}
\usepackage{amsmath}
\usepackage{makecell}
\usepackage{rotating}
\usepackage{booktabs}
\usepackage{xcolor}
\usepackage{placeins}

\journal{Pattern Recognition}

\usepackage{amssymb}% http://ctan.org/pkg/amssymb
\usepackage{pifont}% http://ctan.org/pkg/pifont
\newcommand{\cmark}{\ding{51}}%
\newcommand{\xmark}{\ding{55}}%

\begin{document}

\begin{frontmatter}



\title{Single Image, Any Face: Generalisable 3D Face Generation}


\author[label1]{Wenqing Wang}
\author[label1]{Haosen Yang}
\author[label1]{Josef Kittler}
\author[label1]{Xiatian Zhu \footnote{Corresponding author.}}

\affiliation[label1]{organization={Centre for Vision, Speech and Signal Processing, University of Surrey},%Department and Organization
            % addressline={Guildford, Surrey GU2 7XH, United Kingdom}, 
            % city={Guildford},
            % postcode={GU2 7XH}, 
            % state={Guilford},
            country={United Kingdom}}

% \cortext[cor1]{Xiatian Zhu. Email: xiatian.zhu@surrey.ac.uk}

%% Abstract
\begin{abstract}
The creation of 3D human face avatars from a single unconstrained image is a fundamental task that underlies numerous real-world vision and graphics applications. Despite the significant progress made in generative models, existing methods are either less suited in design for human faces or fail to generalise from the restrictive training domain to unconstrained facial images. To address these limitations, we propose a novel model, {\bf Gen3D-Face}, which generates 3D human faces with unconstrained single image input within a multi-view consistent diffusion framework. Given a specific input image, our model first produces multi-view images, followed by neural surface construction. 
% To incorporate face geometry information in a generalisable manner, we utilise input-conditioned mesh estimation instead of a ground-truth mesh along with the synthetic multi-view training data. 
To incorporate face geometry information while preserving generalisation to in-the-wild inputs, we estimate a subject-specific mesh directly from the input image, enabling training and evaluation without ground-truth 3D supervision.
Importantly, we introduce a multi-view joint generation scheme to enhance the appearance consistency among different views. To the best of our knowledge, this is the first attempt and benchmark for creating photorealistic 3D human face avatars from single images for generic human subject across domains. Extensive experiments demonstrate the efficacy and superiority of our method over previous alternatives for out-of-domain single image 3D face generation and the top ranking competition for the in-domain setting. Our code and dataset will be released upon acceptance.
\end{abstract}

% not claim diffusion-based and triplane-GAN based which better, just claim difference

\begin{graphicalabstract}
The creation of 3D human face avatars from a single unconstrained image is a fundamental task that underlies numerous real-world vision and graphics applications. Despite the significant progress made in generative models, existing methods are either less suited in design for human faces or fail to generalise from the restrictive training domain to unconstrained facial images. To address these limitations, we propose a novel model, {\bf Gen3D-Face}, which generates 3D human faces with unconstrained single image input within a multi-view consistent diffusion framework. Given a specific input image, our model first produces multi-view images, followed by neural surface construction. 
% To incorporate face geometry information in a generalisable manner, we utilise  input-conditioned mesh estimation instead of a ground-truth mesh along with the synthetic multi-view training data. 
To incorporate face geometry information while preserving generalisation to in-the-wild inputs, we estimate a subject-specific mesh directly from the input image, enabling training and evaluation without ground-truth 3D supervision.
Importantly, we introduce a multi-view joint generation scheme to enhance the appearance consistency among different views. To the best of our knowledge, this is the first attempt and benchmark for creating photorealistic 3D human face avatars from single images for generic human subject across domains. Extensive experiments demonstrate the superiority of our method over previous alternatives for out-of-domain single image 3D face generation and the top ranking competition for the in-domain setting.

\end{graphicalabstract}

\begin{highlights}
\item We investigate the under-studied single-image 3D face generation problem with a particular focus on the developed model ability to generalise to unconstrained unseen face imagery so that it is more practically useful and deployable. To the best of our knowledge, this is the very first attempt at tackling this meaningful problem setting in the single image 3D face generation framework with multi-view diffusion model.
\item We propose Gen3D-Face, a novel framework that combines input-conditioned mesh estimation (providing geometric guidance without requiring ground-truth meshes) and joint multi-view diffusion generation (ensuring identity-preserving cross-view consistency), supported by a hybrid real-and-synthetic training strategy that improves robustness in out-of-domain scenarios.
\item Extensive evaluations demonstrate that our method achieves state-of-the-art performance while offering practical advantages in scalability and applicability to unconstrained real-world single-image inputs, since it removes the need for specialised multi-view capture or ground-truth mesh supervision.

% Fill everything
% Figure 2, text more big, subset -> views
% short name for table
% remove and
% separate ablation table, give detail capture for each
% make the cite can link
\end{highlights}

%% Keywords

\end{frontmatter}



% \section{cover letter}
% \input{doc/_cover_letter}

% \section{author bio}
% \input{doc/_author_bio}

\section{Introduction}
\label{sec:intro}
%% Labels are used to cross-reference an item using \ref command.

\begin{figure*}[t]
    \begin{center}
    % \fbox{\rule{0pt}{2in} \rule{.9\linewidth}{0pt}}{}
    \includegraphics[width=.99\linewidth]{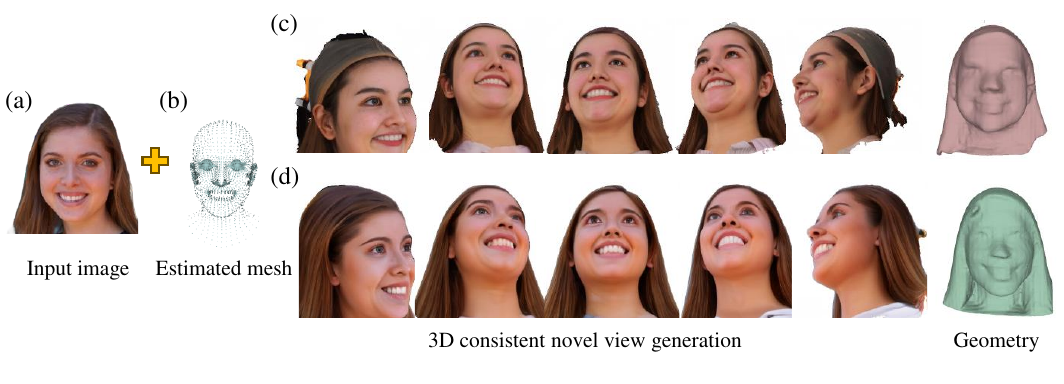}
    \end{center}
    % \vspace{-0.8cm}
   \caption{
   3D human face avatar from (a) a single unconstrained image by (c) prior state-of-the-art model \cite{chen2024morphable} (note the hallucinated hat and clear identity shift), vs. (d) our model.
   }
\label{fig:display}
\end{figure*}

Generate photorealistic 3D face avatars is crucial for a wide range of real-world applications in computer graphics and computer vision, including video conferencing, virtual modeling, entertainment, and augmented or enhanced reality \cite{trevithick2023, huang2024efficient, zeng2022realistic}. Existing 3D face modeling approaches predominantly rely on either multi-view image or video inputs \cite{hong2022headnerf, xu2023deformable, zielonka2023instant, qian2024gaussianavatars} or text-guided 3D generation \cite{zhang2023dreamface, han2023headsculpt}, each presenting inherent limitations. Multi-view reconstruction methods \cite{hong2022headnerf, xu2023deformable, zielonka2023instant, qian2024gaussianavatars} require calibrated multi-view captures and often involve costly per-identity optimization, restricting their applicability in practical scenarios where such input data is unavailable. On the other hand, text-guided avatar generation \cite{zhang2023dreamface, han2023headsculpt} struggles to ensure identity authenticity and preserve fine-grained facial details, as human identity characteristics are difficult to accurately express through text descriptions alone. These constraints collectively highlight the need for a more accessible and identity-faithful solution to 3D face avatar generation.
% The capability to generate photorealistic 3D face avatars from a single image input is essential for a wide range of real applications in computer graphics and computer vision, e.g., video conferencing, virtual modeling, entertainment, augmented and enhanced reality \cite{trevithick2023, huang2024efficient, zeng2022realistic}. 
% The majority of existing 3D face modelling methods not only need costly per-identity optimisation, but also demand input in the form of short text description \cite{han2023headsculpt}, or multi-view images or videos \cite{qian2024gaussianavatars}. 
% Text-guided 3D avatar generation \cite{han2023headsculpt} often struggles to ensure authenticity and identity control, as it faces the daunting task of accurately capturing human identity and face appearance in high detail, unlike image/video-based approaches.
% On the other hand, the latter \cite{ qian2024gaussianavatars} typically rely on multiple view calibrated images, making them less useful and applicable in practice as in many situations such input data is unavailable.

Inspired by the remarkable success of generative diffusion models \cite{rombach2022high} and driven by the aforementioned challenges, {\em single-image 3D face generation} has become a trendy topic with the key challenges being the tasks of figuring out both geometry and appearance information from only a single face image of a generic human identity.
These seemingly impossible tasks now become hopeful for two reasons:
{\em The first} lies in the availability of unprecedentedly rich and comprehensive knowledge captured by off-the-shelf generative models, providing a chance of extracting and transferring useful information for particular downstream tasks (human face in this work) \cite{poole2022dreamfusion, liu2023zero}.
For example, Stable Diffusion was trained with a massive (unknown) text-image pairs from the Internet, including a diversity of facial images from a broad range of subjects like celebrities.
{\em The second} is the enormous technical advance in multi-view image generation \cite{shi2023MVDream, liu2023syncdreamer}, 3D object representation, reconstruction, and generation \cite{mildenhall2021nerf, wang2023neus2, kerbl20233d}.
Combining all these building blocks together properly could be the basis of plausible solutions to tackling this challenge.

Building on the pillars discussed above, an intuitive approach is to learn a generic 3D face generation model from a large, diverse collection of data with multi-view images per human identity, so that the model could generalize to generic unseen single face images.
There are some early attempts pursuing this strategy by training on large synthetic digital avatars created by 3D artists \cite{wang2023rodin}.
This however raises the synthetic to real domain generalisation challenge,
resulting in unrealistic face generation.
Besides, the collection of human face data is much more restricted, due to both the intrinsic complexity and diversity, as well as the intricate privacy considerations.
As a result, existing 3D face benchmarks are often limited in size and diversity in practice, e.g., containing only a few hundred identities \cite{yang2020facescape, pan2024renderme}, making them insufficient for model training.

To mitigate this data scarcity challenge, the latest attempt for single-image 3D face generation leverages the human geometric priors by incorporating ground-truth mesh in multi-view synthesis \cite{chen2024morphable}. A promising finding from this work is that properly blending image appearance and mesh's geometric knowledge enables the model to work across different views, producing  good quality outputs. 
However, we find that their method suffers from several limitations that significantly hamper its generalisation to unconstrained face images shown in Figure~\ref{fig:display}:
(i) {\em Overfitting to the training domain} due to the stringent need for training data. The limited data availability prevents the model to generalise to different unseen styles;
(ii) {\em Over reliance on the ground-truth mesh}, which is often unavailable in practice; 
(iii) {\em Insufficient multi-view consistency} because of multi-view information does not communicate inside Unet Encoder.

% because of the 
% diffusion with  volumes are introduced as condition to maintain the diffusion model's multi-view consistency is lack of 

% its single-view generation design per round.

% In this work, to overcome these limitations
% we propose a novel diffusion-based generative approach, {\bf Gen3D-Face}, for more generalisable 3D face generation using unconstrained single images. 
In this work, to tackle this objective, we develop {\bf Gen3D-Face}, a latent diffusion framework that jointly generates multi-view consistent images and leverages input-estimated face geometry along with hybrid real-and-synthetic training data to reconstruct photorealistic and generalisable 3D face avatars from a single unconstrained image.
Our model first generates consistent multi-view face images and then conducts the neural surface construction. 
To enhance the data diversity, we generate synthetic 3D face images with off-the-shelf model \cite{An_2023_CVPR}.
% Instead of requiring a ground-truth mesh, 
Unlike prior diffusion-based avatar models that rely on ground-truth 3D head scans, which are expensive to obtain and raise privacy concerns,
we exploit input-conditioned mesh estimation for not only mitigating the model's over reliance on the geometric prior, but also enabling it to generalise to typical cases without the ground-truth mesh, and with distinct appearance styles. 
% To more effectively leverage synthetic and real data jointly, 
To ensure multi-view consistency, we introduce a multi-view joint generation scheme. 
% \wwq{The motivation for using  multi-view self attention is the required synthesis dataset quality, which needs to be more effective to improve multi-view consistency}

% While for arbitrary objects, it's generated multi-view images has a certain consistency fault-tolerant, and for Facescape, it has high quality consistency images and 20 expressions for same person, therefore can get good consistency through this,

Our {\bf contributions} are summarised as follows:
{\bf(1)} We investigate the under-studied single-image 3D face generation problem with a particular focus on the developed model ability to generalise to unconstrained unseen face imagery so that it is more practically useful and deployable. To the best of our knowledge, this is the very first attempt at tackling this meaningful problem setting in the single image 3D face generation framework  with multi-view diffusion model.
% {\bf(2)} We propose a novel approach, Gen3D-Face, characterised by the generalisable incorporation of face geometric priors, multi-view joint generation, and joint mining of both real and synthetic 3D face data.
% {\bf(3)} An extensive evaluation of the proposed generalised single image 3D face generation method is carried out. The results demonstrate its superior performance over the state-of-the-art alternatives.
{\bf(2)} We propose Gen3D-Face, a novel framework that combines input-conditioned mesh estimation (providing geometric guidance without requiring ground-truth meshes) and joint multi-view diffusion generation (ensuring identity-preserving cross-view consistency), supported by a hybrid real-and-synthetic training strategy that improves robustness in out-of-domain scenarios.
{\bf(3)} Extensive evaluations demonstrate that our method achieves state-of-the-art performance while offering practical advantages in scalability and applicability to unconstrained real-world single-image inputs, since it removes the need for specialised multi-view capture or ground-truth mesh supervision.

\begin{figure*}[t]
\begin{center}
% \fbox{\rule{0pt}{2in} \rule{.9\linewidth}{0pt}}
\includegraphics[width=1\linewidth]{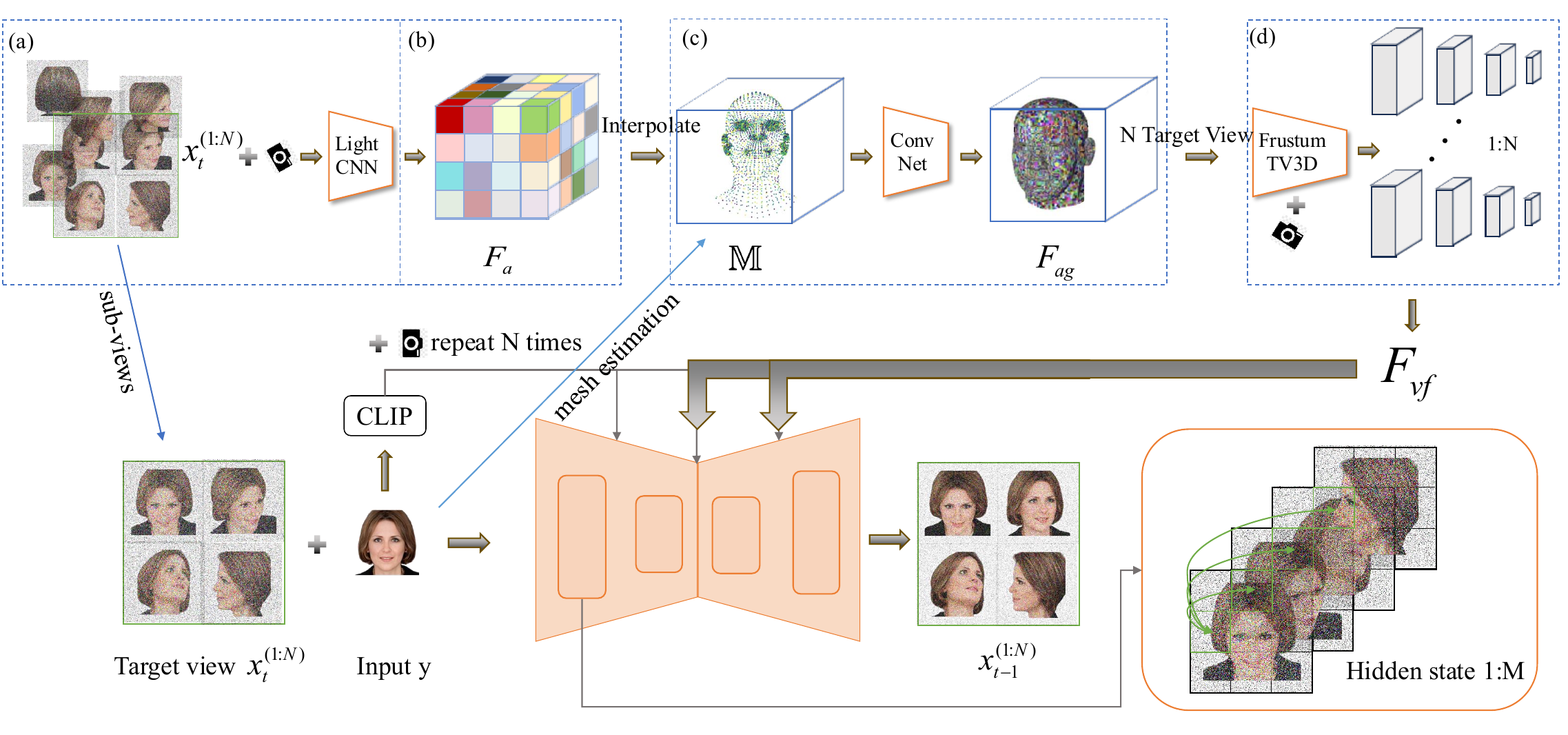}
\end{center}
   \caption{{\bf An overview of our Gen3D-Face}. It adopts the latent diffusion paradigm involving the learning of multi-step denoising. 
   Each step denoises $N$ novel views conditioned on a single face image $y$ and the mesh $\mathbb{M}$ estimated from $y$,
   following the process outlined as below:
   % Taking a single face image $y$ as input, 
   % we start with estimating the mesh $\mathbb{M}$ from image $y$ to impost facial geometry prior.
   % $\mathbf{x}^{(1:N)}_t$ corresponding multi-view images which added noise with time step t, from our synthesis dataset and Facescape. 
   (a) A light CNN encoder is used to integrate the  noise multi-view images $\mathbf{x}^{(1:N)}_t$ generated in the previous steps with camera angles and time embedding;
   (b) Its output is interpolated with a predefined 3D voxel to obtain the {\em appearance feature volume} $F_{a}$; 
   (c) Combining $F_{a}$ with the geometry prior $\mathbb{M}$ yields the {\em hybrid feature volume} $F_{ag}$;
   (d) 
   % Finally obtaining the denoised views $\mathbf{x}^{(1:N)}_{t-1}$ by injecting $F_{ag}$
   % into the diffusion backbone as the condition.
   Finally, the denoised views $\mathbf{x}^{(1:N)}_{t-1}$ are obtained by injecting $F_{ag}$ to FrustumTV3DNet to obtain view frustum volume $F_{vf}$, which is fed into the diffusion backbone as the conditioning signal. 
   Additionally, a CLIP-extracted global feature of the input image $y$ is fused with the camera embedding of each target view, yielding $N$ view-specific conditioning vectors that guide the multi-view diffusion process.}
   % the output of the multi-view UNet is used to denoise $\mathbf{x}^{(1:N)}_t$ to obtain  $\mathbf{x}^{(1:N)}_{t-1}$.
   % This figure gives an overview of a single denoising step of the proposed pipeline. Our model takes a single image $y$, a estimated mesh vertex $\mathbb{M}_{v}$, and corresponding multi-view images $x_{t}^{1:N}$ 
   % (It is latent spaces go through VAE encoder, expressed as images for brevity) which added noise with time step t, from our synthesis dataset and Facescape. 
   % The up
   % The multi-view images $x_{t}^{1:N}$ combined with camera parameters after a encoder, get the 3D Voxel,      Finally, the output of the multi-view UNet is used to denoise $x_{t}^{1:M}$ to obtain $x_{t-1}^{1:M}$
\label{fig:pipeline}
\end{figure*}

\section{Related Work}
\label{sec:related_work}

\noindent \textbf{Novel view synthesis} 
Neural fields \cite{mildenhall2021nerf} and 3D Gaussian Splatting \cite{kerbl20233d} have emerged as the most effective 3D object and scene representations, capable of producing photorealistic images from arbitrary novel views of a scene. However, the first generation is reconstruction-based, necessitating densely captured views. 
To relax this assumption, follow-up approaches \cite{yu2021pixelnerf, szymanowicz2024splatter} propose learning-based methods that require only a few views, utilizing scene priors from other existing datasets \cite{yu2021pixelnerf}, or explicitly mapping the input image to a 3D Gaussian per pixel \cite{szymanowicz2024splatter}. 
Commonly, these methods tend to be restricted to reconstructing relatively simple objects or confined to low resolution, due to their limited expressive capacity.

\noindent \textbf{3D avatars from a single image} 
In addition to reconstruction techniques, various methods have been developed to generate 3D avatars using Generative Adversarial Networks (GANs) \cite{ chan2022efficient, An_2023_CVPR} or, more recently, diffusion models \cite{rombach2022high}.
3D-aware GANs learn 3D representation by integrating tri-planes \cite{chan2022efficient} or tri-grids \cite{An_2023_CVPR} combined with the camera position. 
To achieve a single image 3D avatar generation, typically, GAN inversion is required to fit the input image, which is computationally expensive and time-consuming. 
Live3D \cite{trevithick2023} trains an image-to-triplane encoder to map an input image to a canonical triplane 3D representation instead of GAN inversion, while being still limited to large output angles.
On the other hand, diffusion methods specifically designed for human avatars suffer from limited training data, as a 3D diffusion model is hard to learn from 2D image collections. 
For example, Morphable Diffusion \cite{chen2024morphable} relies on pretrained 3D-aware diffusion models \cite{rombach2022high, liu2023zero} and incorporates 3D physical constraints \cite{loper2023smpl} (ground-truth bilinear head meshes), while being trained on a relatively small and in-domain dataset. These restrictions limit its ability to generalise to unconstrained face images and requires specialised 3D acquisition.
% Therefore, these methods rely on pretrained models \cite{rombach2022high, liu2023zero} and incorporate 3D physical constraints \cite{loper2023smpl} as prior knowledge. However, their stringent input requirements significantly restrict their ability to generalise across out-of-domain face images and situations without ground-truth mesh. 
In this work, we tackle these challenges with a proper model design and data synthesis.

\noindent \textbf{Multi-view diffusion models}
Recent works \cite{liu2023zero, liu2023syncdreamer, shi2023MVDream} extend 2D diffusion models to generate consistent multi-view images from a single-view. 
Their success benefits from the existence of large-scale 3D datasets \cite{deitke2024objaverse}. 
For example, SyncDreamer [11], is a generic multi-view diffusion model trained on large 3D object datasets without a face-specific geometric prior, leading to weaker identity preservation on human faces.
% These multi-view diffusion models designed for arbitrary objects require highly generalization, but have certain fault-tolerant. 
Extending along this direction, our work focuses on human face avatar generation with a special requirement on the model's ability to generalise to unconstrained imagery.

% can be considered as 3D consistent multi-view diffusion model specifically designed for human avatar. 
 % We limit the object to the human head, but require higher consistency, both between the generated images, and between the input and output.

\noindent \textbf{Learning from synthetic data}
Synthetic photorealistic data is effective in handling data scarcity. Recent methods have been developed to utilise synthetic data, either explicitly \cite{lan2023gaussian3diff, deng2024portrait4d} or implicitly \cite{trevithick2023}, to enhance their performance in generative tasks. Portrait4D \cite{deng2024portrait4d} focus more on motion-driven reenactment in the limited poses. The Gaussian Splatting method \cite{lan2023gaussian3diff} needs per-identity optimization. In this work, we extend and validate this generic idea for the more challenging single image 3D face generation in unconstrained settings for a full $360^\circ$.

% \cite{lan2023gaussian3diff} utilize an unconditional 3D GAN \cite{An_2023_CVPR} to generate unconditional head avatar, which needs image inversion to mirror the input image. \cite{trevithick2023} distill the knowledge from a pretrained 3D GAN into a feedforward encoder, but it struggles to deal with a strong profile view because of it is out-of-domain. 

\section{Method}
\label{sec:method}
Given a single face image $y$ as input,
we aim to generate a 3D face avatar for this person.
To that end,
we propose a new latent diffusion approach, {\bf Gen3D-Face},
with the architecture depicted in Figure~\ref{fig:pipeline}.
It generates consistent multi-view images from a single face image,
which can then be fed into existing neural surface construction methods (e.g., Neus2\cite{wang2023neus2}).
For the former, we adopt the off-the-shelf Stable Diffusion \cite{rombach2022high} as the backbone,
where the diffusion and denoising take place in a latent feature embedding space (e.g., a pretrained VAE).
For the sake of being self-contained, we first briefly describe 2D and 3D diffusion.

\subsection{Preliminaries: 2D and 3D Diffusion}
\label{sec:preli}

Diffusion models \cite{rombach2022high} aim to gradually generate structured outputs of a target distribution from random noise through learning an iterative denoising model. 
Given a noise input $x_{t}$, where $t\in (0,T)$ denotes the step index with a total number of steps $T$,
the model is trained to predict the added noise. If this noise is removed, a less noisy version $x_{t-1}$ can be unveiled.
% Given a sample $x_{0}$ following the distribution $P_{x}$, the training objective is to predict noise added by the forward process, denoted as $\epsilon_{\theta}$, to recover $x_{t-1}$ from $x_{t}$, where $t\in (0,T)$ denotes the time. 
Whilst these models can generate images of novel views, it has been demonstrated that it is hard to maintain multi-view consistency \cite{liu2023zero}.

To address this issue, multi-view diffusion has been recently developed \cite{liu2023syncdreamer, shi2023MVDream}.
The key idea is jointly to denoise the images for multiple predefined viewpoints conditioned on the same input $y$,
so that a conditional joint distribution of all these views $p_{\theta}( x^{(1)}_{0}, \cdots ,x^{(N)}_{0}|y)$ can be learned instead, where $N$ specifies the view number.
The forward process adds the same noise to every viewpoint independently at time $t$, and the reverse process is constructed as:
% but  invariably find it difficult to maintain multi-view consistency across different views. This has motivated the current interest in multi-view diffusion \cite{liu2023syncdreamer, shi2023MVDream}.

% Multi-view diffusion aims  to generate N images of predefined viewpoints $\{ x^{(i)}_{0} \}^{N}_{i=1}$, where suffix i denotes the i-th viewpoint and 0 means the time step, by learning a joint distribution of all these views conditioned on a input image $y$: $p_{\theta}( x^{(1)}_{0}, \cdots ,x^{(N)}_{0}|y)$. The forward process adds noise to every view point independently, and the reverse process is constructed as: 
\begin{equation}
    p_\theta(\mathbf{x}_{0:T}^{(1:N)}) 
    =p(\mathbf{x}^{(1:N)}_T) \prod_{t=1}^{T} \prod_{n=1}^{N} p_\theta(\mathbf{x}^{(n)}_{t-1}|\mathbf{x}^{(1:N)}_{t}),
\label{eq:mv_reverse1}
\end{equation}
where the per-step per-view denoising is driven by a Gaussian distribution: 
\begin{equation}
    p_\theta(\mathbf{x}^{(n)}_{t-1}|\mathbf{x}^{(1:N)}_t)=\mathcal{N}(\mathbf{x}^{(n)}_{t-1};\mathbf{\mu}^{(n)}_\theta(\mathbf{x}^{(1:N)}_t,t),\sigma^2_t \mathbf{I}),
\label{eq:mv_p_theta}
\end{equation} 
with the learnable mean for the $n$-th view at step $t$  defined as: 
\begin{equation}
    \mathbf{\mu}^{(n)}_\theta(\mathbf{x}^{(1:N)}_t,t)=\frac{1}{\sqrt{\alpha}_t}\left(\mathbf{x}^{(n)}_t - \frac{\beta_t}{\sqrt{1-\bar{\alpha}_t}} \mathbf{\epsilon}^{(n)}_\theta (\mathbf{x}^{(1:N)}_t, t)\right),
\label{eq:mv_mu1}
\end{equation}
% \begin{equation}
%     \ell=\mathbb{E}_{t,\mathbf{x}^{(1:N)}_0,n,{\epsilon}^{(1:N)}} \left[\|\mathbf{\epsilon}^{(n)} - \mathbf{\epsilon}^{(n)}_\theta (\mathbf{x}^{(1:N)}_t,t)\|_2\right]
% \end{equation}
In equation (\ref{eq:mv_mu1}), $\mathbf{\epsilon}^{(n)}_\theta$ denotes
% are the noise added and 
the trainable noise predictor for the $n$-th view, $\beta_t$, specifies the noise schedule, ${\alpha}_t$ and $\bar{\alpha}_t$ are two scaling constants derived from $\beta_t$.
% $\mathbf{\epsilon}^{(1:N)}$ is the standard Gaussian noise with all $N$ views, 
% and %$\mathbf{\epsilon}^{(n)}$ 
% and 

% \textcolor{red}{alpha and beta undefined}

\subsection{Gen3D-Face}
\label{sec:model}

Extending %the recent attempt
\cite{chen2024morphable} with 
prior multi-view diffusion, 
we take a step towards a generalisable single image 3D face avatar generation, 
where single unconstrained face images are present without ground-truth mesh.
To that end, we first need to address the data scarcity issue as discussed earlier
by multi-view face image synthesis for training data augmentation.
% \noindent {\bf Multi-view face synthesis}
\subsubsection{Synthetic data preprocessing}
\begin{figure*}[t]
\begin{center}
\includegraphics[width=0.99\linewidth]{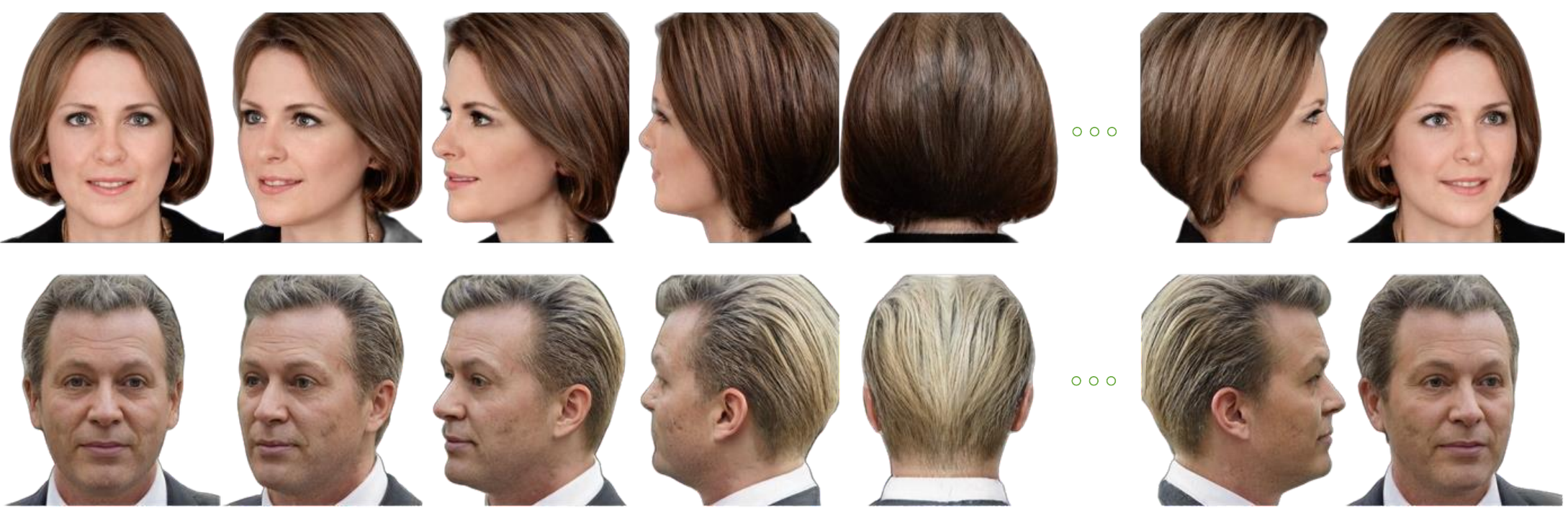}
\end{center}
% \vspace{-0.4cm}
   \caption{Examples of synthetic face images. }
\label{fig:syndata}
\end{figure*}

We adopt the Panohead \cite{An_2023_CVPR} to generate additional training images. 
% This model allows us to generate a large number of virtual human identity face images cross multiple views (see Figure XXX).
We generated 25,000 virtual human identities, each represented by 24 images, with azimuth ranging from -180 to 180 degrees.
% and elevation angle from -40 to 40 degrees
(see Figure~\ref{fig:syndata}).

As the synthesis process is not fully controllable, the output quality is often varying \cite{lan2023gaussian3diff}.
To filter out low-quality face images, we design a pruning process for dealing with the following two issues:
(1) {\em The Janus problem}: We observe cases, where the back-view images present blur faces. To identify such cases, we construct a binary classifier with CLIP using the class names as {\tt back of human head} and {\tt human front face},
and then classify all the back-view face images.
We remove those back-view images with the score of the {\tt human front face} class exceeding a threshold $\tau_{bv}$. 
(2) {\em Identity inconsistency}:
Multi-view face images generated by Panohead \cite{An_2023_CVPR} are likely to be identity inconsistent.
To detect this, we estimate the identity consistency using the average pairwise similarity of views with face embeddings \cite{deng2019arcface} for every individual identity and 
keep only the top-$\tau_{ii}$ virtual identities for model training. 
\subsubsection{Geometry mesh estimation}
To facilitate the 3D face modeling from single images,
we integrate the human head mesh \cite{chen2024morphable} as a prior.
A key difference in our approach is that we use the mesh estimated
from the input image, rather than the ground-truth mesh used in \cite{chen2024morphable}.
The reasons are two-fold: 
(1) Often no ground-truth mesh is available in many real applications;
(2) Using ground-truth mesh tends to make the model over rely on this prior,
whilst largely ignoring the appearance of the input image.
Specifically, we opt for estimating the FLAME mesh $\mathbb{M}$ with $v$ vertices from a single image \cite{RingNet:CVPR:2019, DECA:Siggraph2021, zielonka2022mica} during both training and inference. 
% Also, we use different single image 3D head mesh estimation methods to test if the estimated mesh quality will affect our generation results. 
% As we show in the experiments,
% this design choice is a key to making our model more generalisable. 
This design avoids the need for ground-truth 3D capture, supports privacy-preserving deployment, and as we show in the experiments, effectively prevents the model from overfitting to a fixed mesh template—resulting in better identity preservation and cross-domain generalisation.
We also demonstrate that the influence of different estimation methods \cite{RingNet:CVPR:2019, DECA:Siggraph2021, zielonka2022mica} on the performance can be disregarded.
\subsubsection{Multi-view diffusion generation}
The key in our context is how to effectively condition the multi-view diffusion process with both the appearance of the single image $y$ and the geometry of the estimated mesh  $\mathbb{M}$ (Sec. \ref{sec:preli}).
 
Specifically, let $N$ noisy target views at time $t$ in our multi-view diffusion process be denoted as $\mathbf{x}^{(1:N)}_t$ .
To impose viewpoint information, 
we deploy a CNN encoder to project the camera angles and time embedding to the latent space, which is then added to each novel view's feature embedding $\mathbf{x}^{(1:N)}_t$ respectively. 
To represent these views in the 3D space, 
we construct a 3D volume with its vertex $\mathbb{V} \in \mathbb{R}^{L\times L \times L}$ extracted by a linear sampling along each dimension (where $L$ is the number of voxels in each dimension).
For each novel view $n$, we then warp $\mathbb{V}$ according to this view's extrinsic camera parameters, into which the view's feature embedding $\mathbf{x}^{(n)}_t$ is interpolated.
This results in an {\em appearance feature volume} $F_{a}$ containing $N$ noisy target view features.

To integrate the geometry prior from the estimated mesh $\mathbb{M}$,
we adopt a sparse 3D ConvNet \cite{graham20183d} to interpolate $F_{a}$ with $\mathbb{M}$,
leading to a {\em hybrid feature volume} $F_{ag}$ with both appearance and geometry information. The sparse 3D ConvNet \cite{graham20183d} is a hierarchical CNN network converting the sparse representation to a dense output tensor.
With $F_{ag}$, we produce the {\em view frustum volume} $F_{vf}$ with a light FrustumTV3DNet \cite{liu2023syncdreamer}.  
This $F_{vf}$ serves as a joint condition for multi-view diffusion
by injecting it into the backbone diffusion model (e.g., Stable Diffusion's UNet). The FrustumTV3DNet \cite{liu2023syncdreamer} is a UNet 3D convolutional architecture that processes volumetric input through downsampling stages with time and viewpoint conditioning, followed by skip-connected upsampling paths that return feature maps at multiple resolutions.

As seen from Eq \eqref{eq:mv_p_theta}, the previous methods \cite{chen2024morphable, liu2023syncdreamer} store multi-view information by constructing the 3D volume as a condition. 
% often denoise a single view each time individually, with the condition imposed on the previous step's output of all the views 
% % \textcolor{red}{The meaning not perfectly clear}.
% This design requires $N$ denoising times each for one view, which we consider is inferior in maintaining the view consistency \textcolor{red}{Again, please rephrase.}
To make full use of multi-view information, we propose a {\bf\em multi-view joint generation} algorithm that instead denoises all the views concurrently at one time so that multi-view information interaction can be induced and exploited. 
Specifically, instead of feeding one view $\mathbf{x}^{(n)}_t$ as the decoder's query at a time,
we input sub-views $\mathbf{x}^{(1:N)}_t$ together.
This difference enables us to perform the 3D self attention operation \cite{shi2023MVDream} among all the novel views $\mathbf{x}^{(1:N)}_t$ and the input $y$
for information exchange and to enhance view consistency.

% There is a lack of detailed explanation on how specific models like the light CNN, 3D FrustumTV3D CNN and sparse 3D ConvNet were utilized in this method and what their exact architecture was.

\noindent {\bf Model training}
Our objective function is a multi-view diffusion loss defined as 
\begin{equation}
    \ell(\theta) = \mathbb{E}_{t,y,c,\mathbf{x}^{(1:N)}_0,(1:N),\mathbf{\epsilon}^{(1:N)}} \left[\|\mathbf{\epsilon}^{(1:N)} - \mathbf{\epsilon}^{(1:N)}_\theta (\mathbf{x}^{(1:N)}_t,t)\|_2\right],
\end{equation}
where $y$ is the input image, $c$ represents the camera parameters, $\mathbf{x}^{(1:N)}_0$ denotes the $N$ target-view images, 
$\mathbf{\epsilon}^{(1:N)}$ is the added Gaussian noise, % Multi-view images $x_{1:N}$, 
and $\mathbf{\epsilon}^{(1:N)}_\theta $ is the noise predictor.
% predicted per-view noise by our model. % Unet with multi-self-attention.

\section{Experiments}
\label{sec:experiment}

\paragraph{Datasets}
We adopt the same train/test split protocol as Morphable Diffusion \cite{chen2024morphable}: 323 identities for training, 36 identities for testing, following the official FaceScape \cite{yang2020facescape} evaluation setting.
% For model training, we use the 323 out of 359 identities from the Facescape dataset \cite{yang2020facescape}, following the setting of \cite{chen2024morphable}.
The same real training data is used for all the models compared,
whilst our model also uses synthetic data.
For the {\tt out-of-domain} generalised evaluation, we randomly select 1,024 images from FFHQ \cite{karras2019style} with the background removed using \cite{Qin_2020_PR}. We also test on the H3DS dataset \cite{ramon2021h3d}, which includes multi-view images around the head for 23 identities.
% with 359 subjects, each having 20 unique facial expressions.
% Following \cite{chen2024morphable}, we 
% We train Gen3D-Face on the Facescape dataset \cite{yang2020facescape}, which includes 359 subjects, each having 20 unique facial expressions, and on our synthesis dataset. 
% containing 25,000 identities, with each represented by 48 images, with azimuth ranging from -180 to 180 degrees, and elevation angle from -40 to 40 degrees. 
For the {\tt in-domain} evaluation, as \cite{chen2024morphable} we use the same 36 test identities in the Facescape dataset \cite{yang2020facescape}.

% For the novel view synthesis task, we evaluate performance on both in-domain and out-of-domain datasets. For the in-domain dataset, following the settings in \cite{chen2024morphable}, we divide Facescape into 323 identities for training and 36 identities for evaluation. 
% For out-of-domain evaluation, in line with settings from \cite{trevithick2023, chan2022efficient}, we randomly select 1,024 images from FFHQ \cite{karras2019style}, and remove images background using \cite{Qin_2020_PR}.

\paragraph{Metrics} For the generalised {\tt out-of-domain} evaluation on \textbf{FFHQ} \cite{karras2019style}, where without access to the multi-view images for each identity, we generate 24 views following the test trajectory from Facescape \cite{yang2020facescape}, and evaluate the results using four metrics:
(1) Frechet Inception Distance (FID): calculate FID between all input images with all generated images, 
(2) CLIP Similarity \cite{radford2021learning}: calculate the similarity between the input image and each generated view across all identities,
(3) Input-to-output ID consistency (I2OID): averaging the Arcface cosine similarity \cite{deng2019arcface} between the input image and all generated views, which we propose here to emphasise the importance of  identity preservation,
(4) Output-to-output ID consistency (O2OID): calculated as the mean of Arcface cosine similarity \cite{deng2019arcface} across all pairs of target views generated from the same input image. We repeat this for  the {\tt out-of-domain} dataset \textbf{H3DS} \cite{ramon2021h3d}, which provides captured multi-view images spanning a full $360^\circ$. We generate 24 views for each input image, uniformly spaced at $15^\circ$ intervals from the back to the front. We consider FID, CLIP Similarity and ID consistency between the generated and captured images at matching azimuth angles.
Pixel-level metrics are excluded because the captured images include shoulders, hence the head alignment is inconsistent, and after face cropping, the captured image does not exactly match with the generated images.

For conventional {\tt in-domain} evaluation on \textbf{Facescape}, following \cite{chen2024morphable}, we adopt four metrics: SSIM, LPIPS, FID, and face re-identification accuracy (Re-ID) \cite{phillips2000feret}, calculated between the ground truth and the generated images.
For the Re-ID metric, we consider two variants:
(a) Re-ID(match): As \cite{chen2024morphable}, we calculate the match ratio, which is the percentage of cases, where the Euclidean distance between the generated image and the ground truth image falls below the threshold of 0.6.
% we calculate the percentage of matching the generated image with the ground truth at the Euclidean distance threshold of 0.6;
(b) Re-ID(dist): The average Euclidean distance between the generated images and the ground truth images, in addition to the match ratio, which provides the actual distance value to supplement the matching degree.
% \textcolor{red}{Meaning not clear}.
% We report two face re-identification accuracy values: Re-ID(<0.6) indicates the percentage of cases where, if the Euclidean distance between the vectors of the ground truth and the generated image is less than 0.6, the generated image is classified as re-identifying the same person; and Re-ID is the actual value of the Euclidean distance. 
% In addition, we reconstruct meshes by training the vanilla NeuS for 2k steps, and report the we will show the mesh reconstruction 

\paragraph{Implementation} %Our model is trained with
We use the AdamW optimizer with a batch size of 32 for 90k iterations, training for 4 days on two NVIDIA A100 GPUs (80GB each). 
The learning rate for training the backbone UNet has been raised from 1e-6 to 5e-5 after 100 warm-up steps, and is kept at 5e-4 for all other trainable modules. 
The inference takes about 25 seconds to generate 16 target views from a single input image using 50 DDIM \cite{song2020denoising} steps on an NVIDIA RTX 3090 GPU. 
We set $N=16$ viewpoints, %$M=4$ for multi-view consistency, 
$\tau_{bv}=0.93$ for back-view image filtering, and $\tau_{ii} = 70\%$ for identity consistency filtering.

\paragraph{Competitors} We compare extensively with existing  nerf-based methods, namely pixelNeRF \cite{yu2021pixelnerf}, SSD-NeRF \cite{chen2023single}, and diffusion models including Era3D \cite{li2024era3d}, Zero-1-to-3 \cite{liu2023zero}, SyncDreamer \cite{liu2023syncdreamer}, Morphable Diffusion \cite{chen2024morphable}, and GAN-based methods EG3D \cite{chan2022efficient} and our data generator PanoHead \cite{An_2023_CVPR}. 
Under the proposed {\tt out-of-domain} setting, 
we exclude pixelNeRF \cite{yu2021pixelnerf} and SSD-NeRF \cite{chen2023single}
due to providing no precise camera parameters as required, and
improve the generalisation of Morphable Diffusion \cite{chen2024morphable}
by using the FLAME \cite{li2017learning, zielonka2022mica} meshes obtained by fitting the ground truth 3D keypoints, (originally using ground truth bilinear meshes), otherwise it completely falls apart. All methods are fine-tuned on in-domain training data except Era3D \cite{li2024era3d}, which claims good generalization to human heads, and we do not quantitatively evaluate Era3D \cite{li2024era3d}, as it only generates 6 views.

\begin{table}[tb]
\caption{{\em Out-of-domain} single image 3D face generation results on FFHQ.}
\label{table:out_domain_ffhq}
\begin{center}
% \resizebox{\textwidth}{!}
{
\begin{tabular}{|l|c|c|c|c|}
\hline
Method & \makecell[c]{FID↓} & \makecell[c]{CLIP↑} & \makecell[c]{O2OID↑} & \makecell[c]{I2OID↑} \\
\hline\hline
Zero-1-to-3 \cite{liu2023zero} & 78.8543 & 0.5597  & 0.4483 & 0.1300\\
SyncDreamer \cite{liu2023syncdreamer} & 68.0294 & 0.5983 &  0.4420 & 0.1572 \\
EG3D \cite{chan2022efficient} & 76.1578 & 0.5142 &  0.4623 & 0.1231 \\
PanoHead \cite{An_2023_CVPR} & 58.1578 &  0.6244  & 0.4821 & 0.1611 \\
\hline
Morphable Diffusion \cite{chen2024morphable} & 66.7443 & 0.5959 & \textbf{0.5371} & 0.1596 \\
\bf Gen3D-Face (Ours) & \textbf{54.9575} & \textbf{0.6765} & 0.4936 & \textbf{0.1716} \\
\hline
\end{tabular}
}
\end{center}
\end{table}

\begin{table*}
\caption{{\em Out-of-domain} single image 3D face generation result on H3DS.}
% \vspace{-0.5cm}
\label{table:out_domain_h3ds}
\begin{center}
% \resizebox{\textwidth}{!}
{
\begin{tabular}{|l|c|c|c|}
\hline
Method & FID↓ & CLIP↑ & ID Consistency↑  \\
\hline\hline
Zero-1-to-3 \cite{liu2023zero} & 77.1547 & 0.7612 & 0.1412\\
SyncDreamer \cite{liu2023syncdreamer} & 70.1542 & 0.7814 & 0.1652 \\
EG3D \cite{chan2022efficient} & 180.1254 & 0.6014 & 0.1121  \\
PanoHead \cite{An_2023_CVPR} & 61.2484 & 0.8246 & 0.1811 \\
\hline
Morphable Diffusion \cite{chen2024morphable} & 175.6225 & 0.7493 & 0.0977 \\
\bf Gen3D-Face (Ours) & \bf 59.1536 & \bf 0.8453 & \bf 0.1978  \\
\hline
\end{tabular}
}
\end{center}
\end{table*}

\FloatBarrier  % Forces tables to be placed before the figure

\begin{figure*}[!htbp]
\begin{center}
\includegraphics[width=0.90\linewidth]{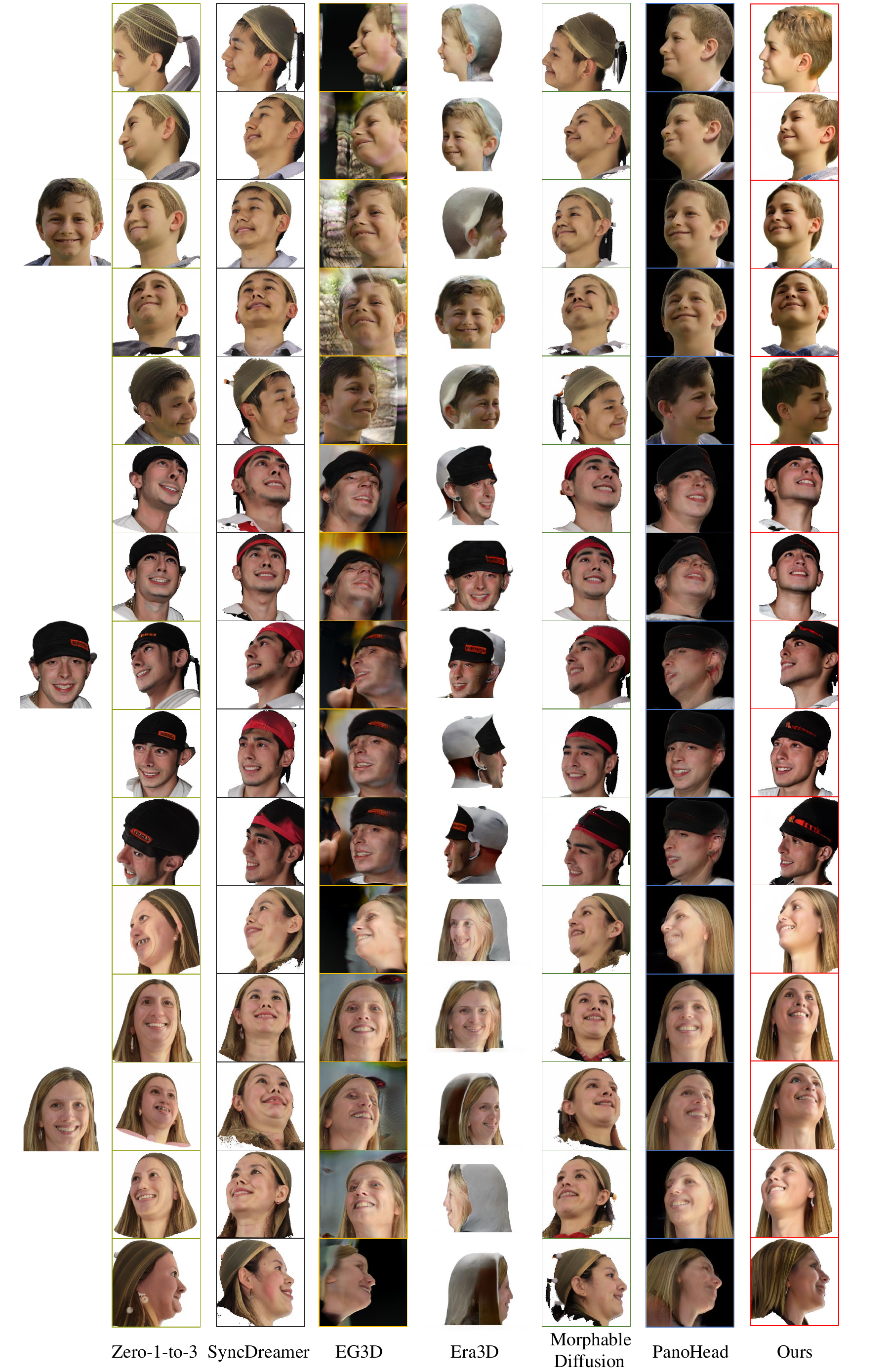}
\end{center}
    \vspace{-0.5cm}
   % \caption{Examples of novel view generation on FFHQ ({\em out-of-domain} setting). The test views come from Facescape \cite{yang2020facescape} testing view except Era3D.}
   \caption{Novel-view generation results in the out-of-domain setting. \textbf{Input}: a single unconstrained FFHQ image (first column). \textbf{Generated target views}: rendered using FaceScape test camera poses (except for Era3D, which can only generate six fixed views). Our method produces more 3D-consistent and identity-preserving results than baselines under large viewpoint changes.}
   % \vspace{-0.2cm}
\label{fig:sotacompare}
\end{figure*}

\begin{figure*}[ht!]
\begin{center}
\includegraphics[width=0.99999\linewidth]{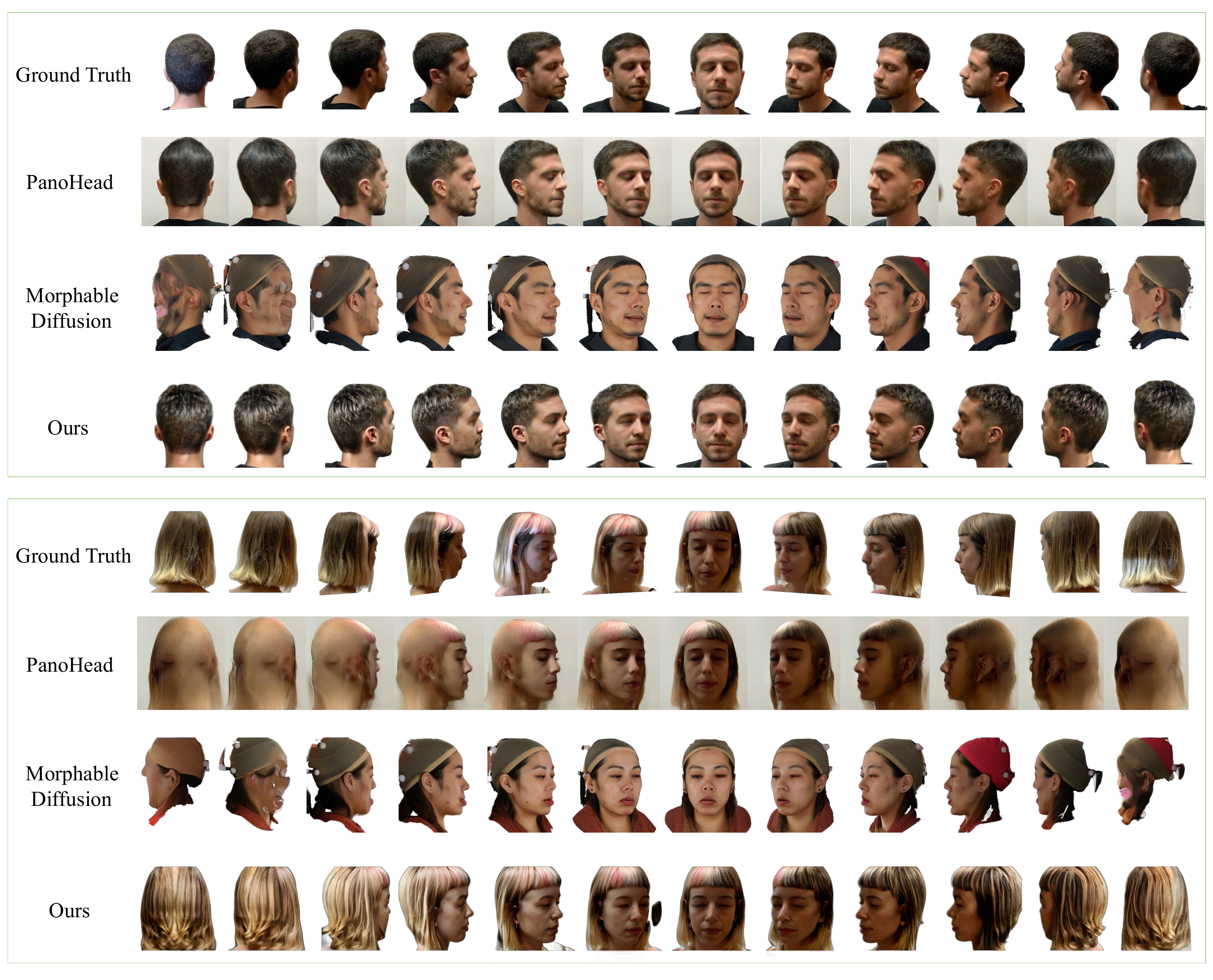}
\end{center}
    \vspace{-0.5cm}
   % \caption{Examples of novel view generation for the H3DS ({\em out-of-domain} setting). The test views are uniformly sampled across $360^\circ$. }
    \caption{Novel-view reconstruction on H3DS ({\em out-of-domain setting}). 
    \textbf{Input}: the frontal H3DS image (middle column). 
    \textbf{Generated views}: multi-pose renderings compared against ground-truth views uniformly sampled across $360^\circ$. 
    Our method better preserves facial identity and geometric structure across large viewpoint variations than prior methods.}
\label{fig:sotacompare_h3ds}
\end{figure*}

\subsection{Evaluation}
\paragraph{Out-of-domain evaluation}
From the \textbf{quantitative} results in Table ~\ref{table:out_domain_ffhq} and Table ~\ref{table:out_domain_h3ds}, 
we observe that:
{\bf (1)} Interestingly, the diffusion model with head geometry guidance \cite{chen2024morphable} does not outperform generic object diffusion models (Zero-1-to-3 \cite{liu2023zero}, SyncDreamer \cite{liu2023syncdreamer}) after fine-tuning, nor earlier GAN models (PanoHead \cite{An_2023_CVPR}) on three out of four metrics in unseen domains. 
% Even get worse if the generation spans a full $360^\circ$ images (Table ~\ref{table:out_domain_h3ds}). 
Its performance further degrades when the generation spans a full $360^\circ$ range of views (Table ~\ref{table:out_domain_h3ds}).
This suggests that the effectiveness of imposing human geometry in a limited size is constrained. In contrast, our proposed synthetic dataset can improve this limitation.
% This is {\em contrary} to what was discovered in \cite{chen2024morphable}, under this more challenging setting,
% suggesting that the evaluation setting {\em matters}, fundamentally.
% This also implies that imposing human geometry to improve performance in the limited in-domain setting, as in \cite{chen2024morphable}, is directly responsible for the degraded generalisation.
{\bf (2)} The GAN model used to create our synthetic dataset achieves the second-best performance when generating multi-view images without large elevation angles (Table ~\ref{table:out_domain_h3ds}). However, its advantage becomes less clear when views include larger elevation angles, as shown in Table~\ref{table:out_domain_ffhq}. Note that EG3D \cite{chan2022efficient} yields almost the worst results because its mainly designed for limited views.
{\bf (3)} Overall our Gen3D-Face is the best performer, except being second to Morphable Diffusion \cite{chen2024morphable} on the output-to-output ID consistency metric. We note, however,  that looking at this metric {\em alone} is not comprehensive, and even misleading, since it overlooks the divergence of the generated images from the input (e.g., being consistent multi-view images of a totally different identity).
Instead, we should jointly consider both input-to-output and output-to-output ID consistency.
% Fusing the two metrics could make the comparison easier but hard to make sense out of it.

\begin{figure}[!h]
\begin{center}
\includegraphics[width=0.99999\linewidth]{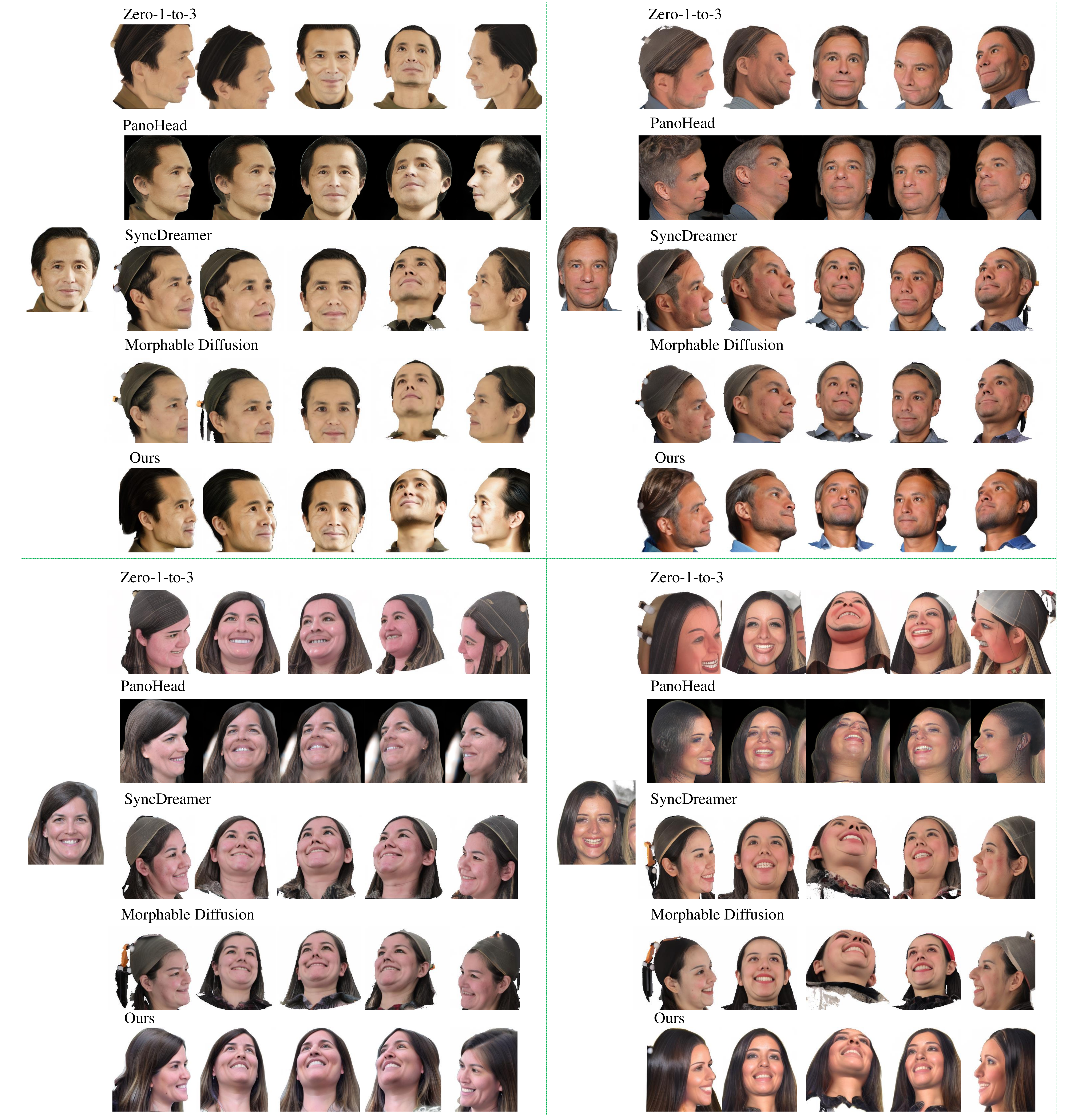}
\end{center}
   \vspace{-0.4cm}
   \caption{More examples of the
novel-view generation results in the out-of-domain setting. Input: a single unconstrained FFHQ
image (first column). Generated target views: rendered using FaceScape test camera poses. Our method produces more 3D-consistent and identity-
preserving results than baselines under large viewpoint changes. }
\label{fig:out_of_domain}
\end{figure}

The \textbf{qualitative} evaluation is presented in Figure~\ref{fig:sotacompare}, Figure~\ref{fig:sotacompare_h3ds} and Figure~\ref{fig:out_of_domain}, Figure~\ref{fig:sotacompare_h3ds} only contain the best performance methods due to the page limit. We attempt to present consistent camera views across the methods within each row, but slight differences remain due to variations in the training camera parameters across the methods, especially for Era3D \cite{li2024era3d}.
We make these observations:
{\bf (1)} Zero-1-to-3 \cite{liu2023zero} tends to produce cartoon style images;
% {\bf (2)} As Era3D \cite{li2024era3d}, the most recent single-image-to-3D method, can only generate 6 views and visually exhibits unrealistic geometry.
{\bf (2)} Era3D \cite{li2024era3d}, the most recent single-image-to-3D method, can only generate six views and exhibits unrealistic geometry;
{\bf (3)} SyncDreamer \cite{liu2023syncdreamer} and Morphable Diffusion \cite{chen2024morphable} struggle in preserving the identity;
{\bf (4)} Morphable Diffusion  \cite{chen2024morphable} generates images that are more consistent, but it suffers from overfitting to the training domain (e.g., added hat for all cases);
{\bf (5)} EG3D \cite{chan2022efficient} and Panohead \cite{An_2023_CVPR} tends to yield more blurry images, especially the face edge,
despite taking 20$\times $ more training time, which is caused by the PTI inversion \cite{roich2022pivotal};
{\bf (6)} Our Gen3D-Face achieves the overall best result in terms of ID preservation and consistency, and wider pose variation.

\begin{table}[!htbp]
\caption{{\em In-domain} single image 3D face generation result on Facescape.}
\label{table:in_domain}
\begin{center}
% \resizebox{\textwidth}{!}
{
\begin{tabular}{|l|c|c|c|c|c|}
\hline
Method & SSIM↑ & LPIPS↓ & FID↓ & \makecell[c]{Re-ID\\(match)↑} & \makecell[c]{Re-ID\\(dist)↓} \\
\hline\hline
pixelNeRF \cite{yu2021pixelnerf} &  0.7898 & 0.2200 & 92.61 & 0.9746 & 0.3912 \\
Zero-1-to-3 \cite{liu2023zero} & 0.5656 & 0.4248 & 10.97 & 0.9677 & 0.4193 \\
SSD-NeRF \cite{chen2023single} &  0.7225 & 0.2225 &  34.88 & 0.9874 & 0.3855 \\
SyncDreamer \cite{liu2023syncdreamer} & 0.7732 & 0.1854 & \textbf{6.05} &  0.9960 & 0.3391 \\
PanoHead \cite{An_2023_CVPR} & 0.7871 & 0.1914 & 7.10 & 0.9915 & 0.3412 \\
Morphable Diffusion \cite{chen2024morphable} & \textbf{0.8064} & \textbf{0.1653} & 6.73 & \textbf{0.9986} & \textbf{0.3372} \\
\bf Gen3D-Face & 0.7995 & 0.1701 & 6.1231 & 0.9981 & 0.3375\\
\hline
\end{tabular}
}
\end{center}

\end{table}

\begin{figure*}[ht!]
\begin{center}
\includegraphics[width=0.99999\linewidth]{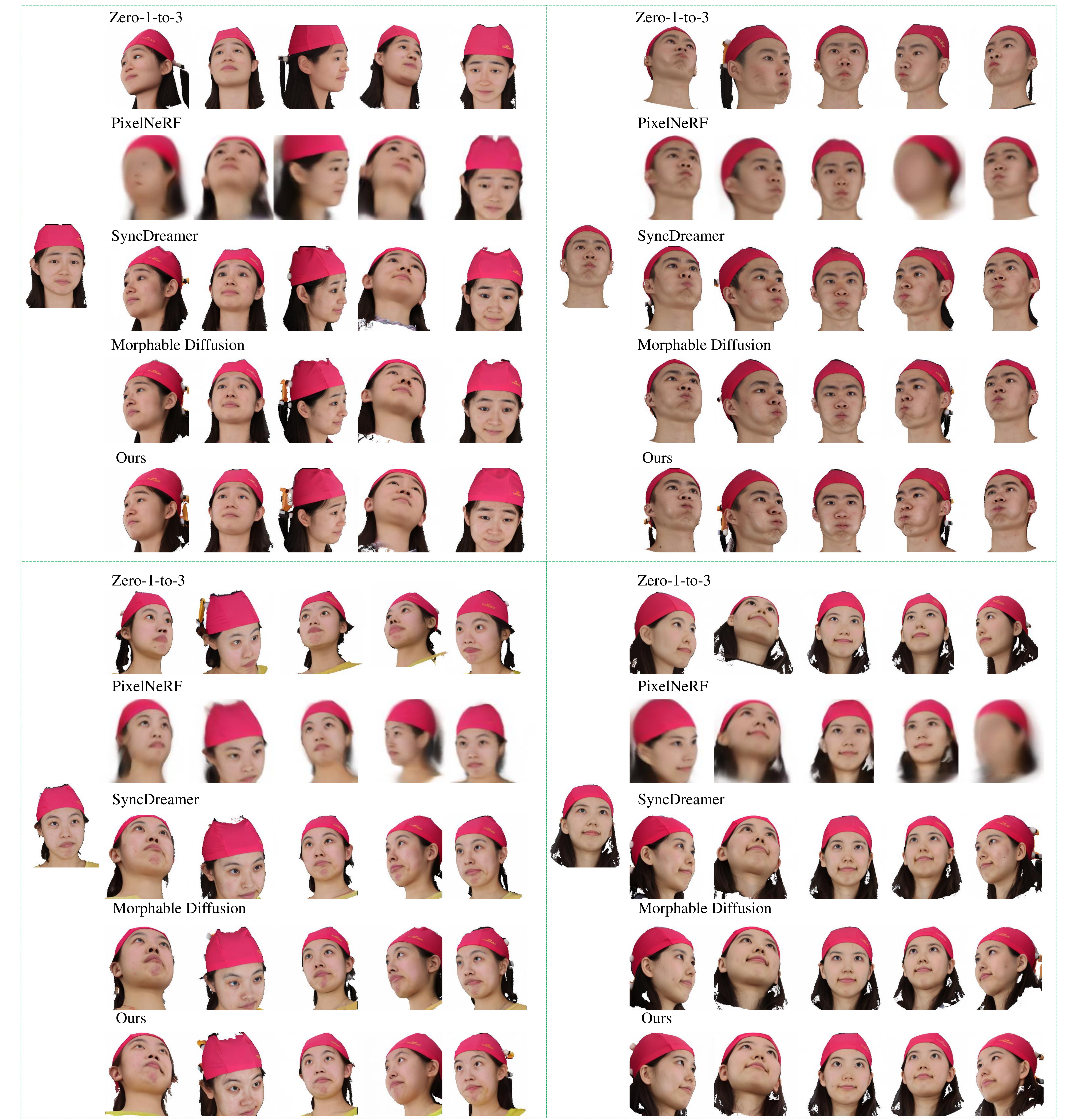}
\end{center}
    % \vspace{-0.4cm}
   % \caption{Examples of novel views generated on the Facescape ({\em in-domain} setting). }
    \caption{Novel-view generation results on FaceScape in the {\em in-domain setting}. 
    \textbf{Input}: a single FaceScape image. 
    \textbf{Generated views}: novel multi-angle renderings compared with those produced by baseline methods under the same viewpoint settings. 
    This suggests that our model does not sacrifice performance in the training domain, while enhancing generalisation ability.
}
    % \vspace{-0.2cm}
\label{fig:in_domain}
\end{figure*}

\paragraph{In-domain evaluation}
While this work stresses the importance of the out-of-domain generalisation,
we still evaluate the conventional in-domain setting.
From Table ~\ref{table:in_domain} we observe that our method performs on par with the previous model, Morphable Diffusion \cite{chen2024morphable}.
This suggests that our model does not sacrifice the training domain performance, while enhancing the model generalisation. The qualitative evaluation in Figure~\ref{fig:in_domain} shows that our method preserves the identity well.
% We provide qualitative evaluation in supplementary file.
% give the results of our method versus other methods for in domain dataset, and the visualization results are in supplementary. Our model fine-tuned on Facescape training set get basically consistent results with morphable diffusion. And we don't use ground truth mesh in both training and fine-tuning process.

\subsection{Ablation studies}
\label{sec:Ablation}

\begin{table}
\caption{Effect of pruning synthetic training data - Filtering out synthetic training data with the Janus problem or identity inconsistencies enhances training performance.
}
% \vspace{-0.5cm}
\begin{center}
\label{table:ablation-data_pruning}
% \resizebox{\textwidth}{!}
{
\begin{tabular}{|c|c|c|c|c|}
\hline
Pruning & \makecell[c]{FID↓} & \makecell[c]{CLIP↑} & \makecell[c]{O2OID↑} & \makecell[c]{I2OID↑} \\
\hline\hline
\xmark & 57.3138  & 0.6624 & 0.4451 & 0.1659 \\
\cmark  & \textbf{54.9575} & \textbf{0.6765} & \textbf{0.4936} & \textbf{0.1716} \\
\hline
\end{tabular}
}
\end{center}
\end{table}

\begin{figure*}
\begin{center}
\includegraphics[width=0.99\linewidth]{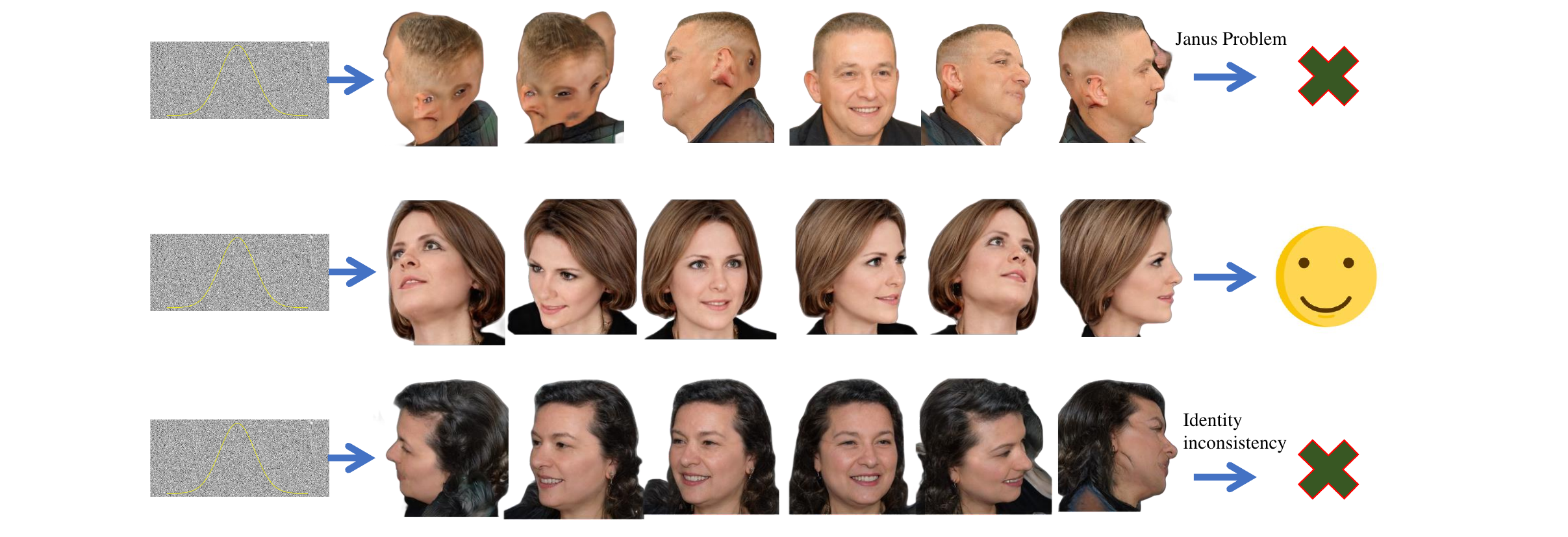}
\end{center}
% \vspace{-0.5cm}
   \caption{The Janus problem and identity inconsistency with the synthetic data.}
% \vspace{-0.5cm}
\label{fig:syn}
\end{figure*}

\noindent \textbf{Data pruning}
We show examples of the Janus problem and Identity inconsistency in Figure~\ref{fig:syn}, which are filtered out using our pruning process. The effect of pruning is shown in Table~\ref{table:ablation-data_pruning}.

\begin{table}[ht]
\caption{
Effect of training data (real, synthetic, or both) on FFHQ.
% \cite{karras2019style}.
% , comparing our model trained on real dataset Facescape \cite{yang2020facescape} only, our synthetic dataset only, and both.
}
% \vspace{-0.5cm}
\begin{center}
\label{table:ablation-real_syn_data}
% \resizebox{\textwidth}{!}
{
\begin{tabular}{|l|c|c|c|c|}
\hline
Training data & \makecell[c]{FID↓} & \makecell[c]{CLIP↑} & \makecell[c]{O2OID↑} & \makecell[c]{I2OID↑} \\
\hline\hline
Real only & 64.2882 & 0.6110 & \textbf{0.5217} & 0.1618 \\
Synthetic only & 85.5801 & 0.5149 & 0.4155 & 0.1011\\ \hline
\bf Both & \textbf{54.9575} & \textbf{0.6765} & 0.4936 & \textbf{0.1716} \\
\hline
\end{tabular}
}
\end{center}
\end{table}

\noindent \textbf{The training data}
We evaluate the effect of synthetic and real training data.
As shown in Table~\ref{table:ablation-real_syn_data},
we find that (1) both real and synthetic data contribute positively, 
but real data is more useful, despite its smaller size. However, the high output-to-output ID consistency suggests overfitting to the training domain.
(2) Relying solely on synthetic data introduces a domain gap when testing on real images.
(3) Using both significantly boosts performance, validating our motivation
for the training data augmentation by synthesis.
% (3) The significant decrease for the exclusive use of synthetic data is because training and evaluation camera views are not the same.

% \input{estimated_mesh}

\begin{table}
\caption{Impact of mesh estimation by different methods on a single image on FFHQ.
We show that this component is not sensitive.
}
% \vspace{-0.5cm}
\begin{center}
\label{table:ablation-estimated_mesh}
% \resizebox{\textwidth}{!}
{
\begin{tabular}{|l|c|c|c|c|}
\hline
Method & \makecell[c]{FID↓} & \makecell[c]{CLIP↑} & \makecell[c]{O2OID↑} & \makecell[c]{I2OID↑} \\
\hline\hline
RingNet \cite{RingNet:CVPR:2019} & 53.4471 & 0.6712 & 0.4912 & 0.1713 \\
DECA \cite{DECA:Siggraph2021}  & 54.9575  & 0.6765 & 0.4936 & 0.1716 \\ 
MICA \cite{zielonka2022mica}  & 54.9812 & 0.6755 & 0.4922 & 0.1716 \\
\hline
\end{tabular}
}
\end{center}
\end{table}

\noindent \textbf{The mesh prior effect}
We evaluate the impact of different mesh priors on the generative model for the following scenarios:
(1) Mesh estimated using RingNet \cite{RingNet:CVPR:2019}
(2) Mesh estimated using DECA \cite{DECA:Siggraph2021}
(3) Mesh estimated using MICA \cite{zielonka2022mica}
As shown in Table~\ref{table:ablation-estimated_mesh}, different methods of estimating the mesh from the input image do not significantly affect the generated results.

\begin{figure*}[!htbp]
\begin{center}
% \fbox{\rule{0pt}{2in} \rule{.9\linewidth}{0pt}}{}
\includegraphics[width=.99\linewidth]{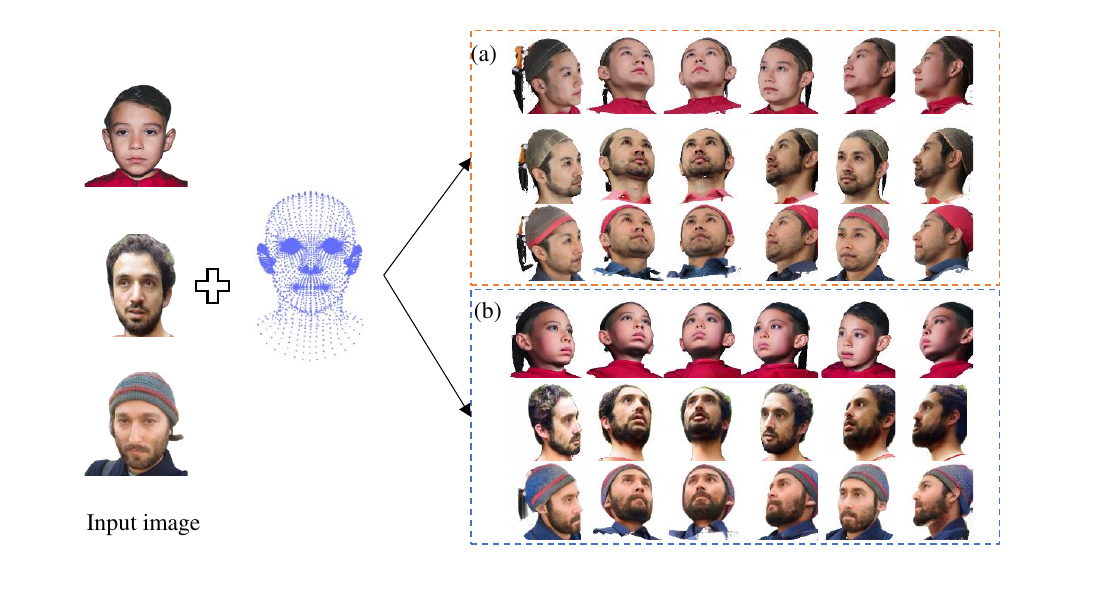}
\end{center}
    % \vspace{-0.4cm}
   \caption{Model trained on different geometry priors: 
   (a) Ground-truth fitted mesh \cite{chen2024morphable}; (b) Our input-estimated mesh. 
   After training, if given different input images with the same mesh, the model trained with (a) tends to generate similar identities despite different inputs.
   % ignore input image in a certain extent, and display similar identity features. 
   In contrast, the model trained on (b) preserves more distinct identity features from each image.
   % more of each image’s unique identity information.
   }
   \vspace{-0.5cm}
\label{fig:estimated_gt_mesh}
\end{figure*}

In Figure~\ref{fig:estimated_gt_mesh}, we also evaluate training with a ground-truth bilinear mesh \cite{chen2024morphable} and the input-estimated Flame mesh \cite{DECA:Siggraph2021} as proposed in our work. We provide different input images and the same mesh (randomly chosen from the Facescape testset), we observed that training with the ground-truth mesh
leads to the challenge of preserving the input identity and appearance information from the image. It will more relying on the input mesh, as we can see in Figure~\ref{fig:estimated_gt_mesh} (a), even input different images, the generated multi-view images will show similar identity features.
In contrast, our solution can let the model also pay attention to the input image, as more specific identity information will come from input image.

\begin{table*}[!htbp]

\caption{%Ablation on the 
The effect of multi-view joint generation (MVJG). 
We increase the batch size (maintaining the same iteration count) which reduces epochs when number of views increase, and the results will decrease as views increase because of under-trains. Although more training might help, we fix four views to manage costs.
}
% \vspace{-0.5cm}
\label{table:mvjg}
\begin{center}
% \resizebox{\textwidth}{!}
{
\begin{tabular}{|l|c|c|c|c|c|}
\hline
Number of views & batch size & \makecell[c]{FID↓}  & \makecell[c]{CLIP↑} & \makecell[c]{O2OID↑} & \makecell[c]{I2OID↑} \\
\hline\hline
\makecell[c]{1} & 70 & 58.4648 & 0.6181 & 0.4441 & 0.1571 \\
\makecell[c]4  & 28 & \bf 54.9575  & \bf 0.6765 & \bf 0.4936 & \bf 0.1716 \\ 
\makecell[c]8 & 8 & 62.1981 & 0.5541 & 0.4122 & 0.1341 \\
\makecell[c]{16} & 4 & 66.4711 & 0.5344 & 0.4013 & 0.1249 \\
\hline
\end{tabular}
}
\end{center}

\end{table*}

\noindent \textbf{Joint multi-view generation}
We evaluate the effect of our joint multi-view generation in Table~\ref{table:mvjg}.
Increasing the views cardinality (as shown in Figure~\ref{fig:pipeline}) requires more training resources due to the attention module. To balance this, we reduce the batch size when the subset number goes up. Because we keep the same number of training iterations across all configurations, higher subset numbers inevitably lead to less complete training. While longer training could improve performance, we set the view cardinality to four to manage the cost of training in all experiments.  

% We set the same training iteration, therefore when subset number increase, the model actually not fully trained, if want to get best performance still need more long training, therefore, think about training cost, we choose 4 in all experiments.

% \begin{figure*}[h]
% \begin{center}
% \includegraphics[width=0.99\linewidth]{comparison_2.pdf}
% \end{center}
%    \caption{Examples of novel view generation on FFHQ ({\em out-of-domain} setting). }
% \label{fig:sotacompare}
% \end{figure*}

% \begin{figure*}
% \begin{center}
% \includegraphics[width=0.95\linewidth]{in_domain.pdf}
% \end{center}
%    \caption{Examples of novel view generation on Facescape ({\em in-domain} setting). }
% \label{fig:in_domain}
% \end{figure*}

% \begin{figure*}
% \begin{center}
% \includegraphics[width=0.99\linewidth]{out_of_domain.pdf}
% \end{center}
%     % \vspace{-0.4cm}

%    \caption{More examples of novel view generation on FFHQ ({\em out-of-domain} setting). }
% \label{fig:out_of_domain}
% \end{figure*}

\section{Limitations}
\label{sec:limitation}
Our work aims to generate multi-view images from any input image and subsequently perform reconstruction. 
% However, due to limitations in the training data, our method is somewhat sensitive to variations in image focal length.
% Unfortunately, obvious occlusions or paintings on the face can also affect the quality of our generation results.
However, the diffusion-based pipeline requires multiple denoising steps per view, resulting in high computation and limiting real-time deployment. Additionally, performance may drop under extremely exaggerated expressions or challenging pose combinations, as input-estimated FLAME meshes and the training data provide limited coverage of such rare geometric deformations, leading to less reliable conditioning.
Also due to the relatively narrow distribution of camera intrinsics in our training data and the current assumption of a fixed focal length in the 3D feature projection, our method can be sensitive to focal-length variations in the input image, which may lead to geometric distortions in novel views. 
Moreover, occlusions or facial decorations (e.g., hands, microphones, large glasses, heavy makeup) remain challenging, as these patterns are under-represented in the training data and can negatively impact mesh estimation and identity embedding reliability.

\section{Conclusion}
\label{sec:conclusion}

In this work, we presented a pioneering investigation of the problem of single image 3D face generation in unconstrained, out-of-domain scenario.
Built on the recent multi-view diffusion approach, 
we proposed a novel generative method, Gen3D-Face, that generates photorealistic 3D human face avatars from single, unconstrained images. 
To achieve this, Gen3D-Face integrates three key components: (i) a hybrid real-and-synthetic training pipeline with quality-controlled augmentation to enhance cross-domain generalisation, (ii) a joint multi-view diffusion scheme that enforces cross-view appearance consistency, and (iii) input-conditioned mesh estimation that provides structural guidance while maintaining flexibility for in-the-wild inputs.
% We showed that the proposed specific design features such as enhanced training data, input-conditioned mesh estimation, and joint multi-view generation are critical to the quality of the generated images. 
% We benchmark this more sophisticated approach with the existing generative methods using comprehensive metrics.
% Extensive experiments show that our method excels in creating unconstrained avatars for generic human subjects, whilst achieving competitive performance under the constrained in-domain setting.
Extensive experiments show that our method not only surpasses state-of-the-art baselines in challenging out-of-domain scenarios, but also remains competitive under standard in-domain settings, demonstrating its strong generalisation capability and broad applicability to real-world single-image 3D avatar generation.
% \textcolor{blue}{Looking forward, we plan to reduce computational cost through accelerated sampling, lightweight backbones, or model distillation, and better handle extreme facial expressions and poses by incorporating expression/pose-conditioned training and more diverse geometric priors.
Looking forward, we plan to reduce computational cost through accelerated sampling, lightweight backbones, or model distillation, enhance robustness to focal-length variation and occlusion through improved camera modelling and occlusion-aware training, and better handle extreme facial expressions and poses by incorporating expression/pose-conditioned training and more diverse geometric priors.

\bibliographystyle{elsarticle-num} 
\bibliography{main}



% \bibliographystyle{unsrt}
% \bibliographystyle{plain}
% \bibliography{main}

\clearpage

\section{Supplementary }

\begin{sidewaystable*}[t]
\centering
\caption{Schematic comparison between our Gen3D-Face and closely related methods.}
\label{tab:method_comparison}

% ===== Block 1 =====
\textbf{(a) Gen3D-Face vs Morphable Diffusion}\\[0.4em]
\begin{tabular}{p{0.18\linewidth}p{0.35\linewidth}p{0.35\linewidth}}
\toprule
 & Gen3D-Face (Ours) & Morphable Diffusion~\cite{chen2024morphable} \\
\midrule
Model type
& Diffusion (latent, multi-view)
& Diffusion (latent, multi-view) \\

Geometry prior
& Input-estimated FLAME mesh
& GT bilinear head mesh \\

Mesh requirement
& No GT mesh; single-image fitting only
& Requires GT mesh for training \\

Training data
& Hybrid real FaceScape + pruned synthetic faces (PanoHead)
& Real FaceScape only (limited domain) \\

Target domain
& 3D human heads (unconstrained)
& 3D human heads (in-domain) \\

Multi-view consistency
& Joint multi-view denoising
& Per-view denoising \\

Identity preservation
& Strong
& Limited, domain overfitting \\

OOD generalisation
& Strong 
& Weak \\
\bottomrule
\end{tabular}

\vspace{1.4em}

% ===== Block 2 =====
\textbf{(b) SyncDreamer vs PanoHead}\\[0.4em]
\begin{tabular}{p{0.18\linewidth}p{0.35\linewidth}p{0.35\linewidth}}
\toprule
 & SyncDreamer~\cite{liu2023syncdreamer} & PanoHead~\cite{An_2023_CVPR} \\
\midrule
Model type
& Diffusion (latent, multi-view)
& GAN (tri-plane) \\

Geometry prior
& None (generic 3D prior)
& None (implicit in tri-plane) \\

Mesh requirement
& No mesh
& No mesh \\

Training data
& Generic object 3D datasets
& Single View datasets (FFHQ) \\

Target domain
& Generic 3D objects
& GAN training domain faces \\

Multi-view consistency
& Synchronized diffusion
& No explicit diffusion coupling \\

Identity preservation
& Weak for faces
& Need GAN inversion \\

OOD generalisation
& Not specialised for faces
& Drops under extreme pose shift \\
\bottomrule
\end{tabular}

\end{sidewaystable*}

\begin{table}[h]
\caption{Combined ablation results on FFHQ showing the effect of synthesis data pruning, input-estimated mesh prior, and multi-view joint generation. Removing any component leads to noticeable degradation across image fidelity, identity similarity, and multi-view consistency metrics.}
\begin{center}
\label{table:ablation-combined}

\begin{tabular}{|l|c|c|c|c|}
\hline
Component & \makecell[c]{FID↓} & \makecell[c]{CLIP↑} & \makecell[c]{O2OID↑} & \makecell[c]{I2OID↑} \\
\hline\hline
w/o synthesis data pruning & 57.3138 & 0.6624 & 0.4451 & 0.1659 \\
w/o mesh prior & 64.0154 & 0.6083 & 0.4320 & 0.1472 \\ 
w/o multi-view joint generation & 58.4648 & 0.6181 & 0.4441 & 0.1571 \\
All & \textbf{54.9575} & \textbf{0.6765} & \textbf{0.4936} & \textbf{0.1716} \\
\hline
\end{tabular}
\end{center}
\end{table}

% \clearpage
% \input{response_R1}


% %% For numbered reference style

%   \textit{\LaTeX: a document preparation system},
%   Addison Wesley, Massachusetts,

% \end{thebibliography}
\end{document}